\documentclass{article}
\usepackage{spconf,amsmath,graphicx}
\usepackage{multirow}
\usepackage{amssymb}
\usepackage{enumitem}
\usepackage{bbding}
\usepackage{hyperref}
\usepackage[ruled]{algorithm2e}

\title{Progressive Meta-Pooling Learning for Lightweight Image Classification Model}

\name{\begin{tabular}{c}{}
Peijie Dong$^{1}$, Xin Niu$^{1,\ast}$\thanks{*Corresponding author}, Zhiliang Tian$^{1}$, Lujun Li$^{2}$, Xiaodong Wang$^{1}$ \\
Zimian Wei$^{1}$, Hengyue Pan$^{1}$, Dongsheng Li$^{1}$
\end{tabular}
}
\address{$^1$ School of Computer, National University of Defense Technology, Changsha 410073, China\\
$^2$ Chinese Academy of Sciences, Beijing, China}

\begin{document}
%
\maketitle
\begin{abstract}

Practical networks for edge devices adopt shallow depth and small convolutional kernels to save memory and computational cost, which leads to a restricted receptive field. Conventional efficient learning methods focus on lightweight convolution designs, ignoring the role of the receptive field in neural network design. In this paper, we propose the Meta-Pooling framework to make the receptive field learnable for a lightweight network, which consists of parameterized pooling-based operations. Specifically, we introduce a parameterized spatial enhancer, which is composed of pooling operations to provide versatile receptive fields for each layer of a lightweight model. Then, we present a Progressive Meta-Pooling Learning (PMPL) strategy for the parameterized spatial enhancer to acquire a suitable receptive field size. The results on the ImageNet dataset demonstrate that MobileNetV2 using Meta-Pooling achieves top1 accuracy of 74.6\%, which outperforms MobileNetV2 by 2.3\%.

\end{abstract}
\section{Introduction}

Lightweight networks tailor specifically for mobile or resource-constrained platforms, with fewer parameters and memory consumption. MobileNet\cite{Sandler2018MobileNetV2IR} and ShuffleNetv2\cite{Ma2018ShuffleNetVP} utilize factorization to make the standard convolution light-weight and computationally friendly. However, the lightweight networks are spatially local, a.k.a. the receptive field is limited.

The receptive field determines how the network aggregates spatial contexts and has a significant impact on the performance of the lightweight model. Evidence shows that a large receptive field is necessary for high-level recognition tasks with diminishing rewards\cite{Liu2022ACF, Ding2022ScalingUY}. Conventional approaches adopt small kernel stacking \cite{SimonyanZ14a,7780459}, pooling-based methods\cite{He2015SpatialPP,Hou2020StripPR,Hou2021CoordinateAF}, and special convolution operations \cite{Dai2017DeformableCN,tan2019mixconv,li2023norm} to enlarge the receptive field. Among them, spatial pooling is playing an important role, which maintains the most important activation for the most distinguishing features. Besides the average pooling and max pooling, recent pooling operations mainly focus on how to preserve the local details when down-sampling the feature map. We explore the potential of pooling operations to aggregate spatial information and expand the receptive field. Other methods \cite{Hu2020SqueezeandExcitationN,Woo2018CBAMCB,Wang2018NonlocalNN} resort to attention mechanism to expand the receptive field from the perspective of attention. Most of the attention-based methods are plugged into a network in a fixed manner and the receptive field is predetermined. However, as revealed in the CoAtNet\cite{dai2021coatnet}, having a larger receptive field does not necessarily means a better performance while having an appropriate local receptive field can improve the generalization ability of the model.

\begin{figure}
  \centering
  \includegraphics[width=0.99\linewidth]{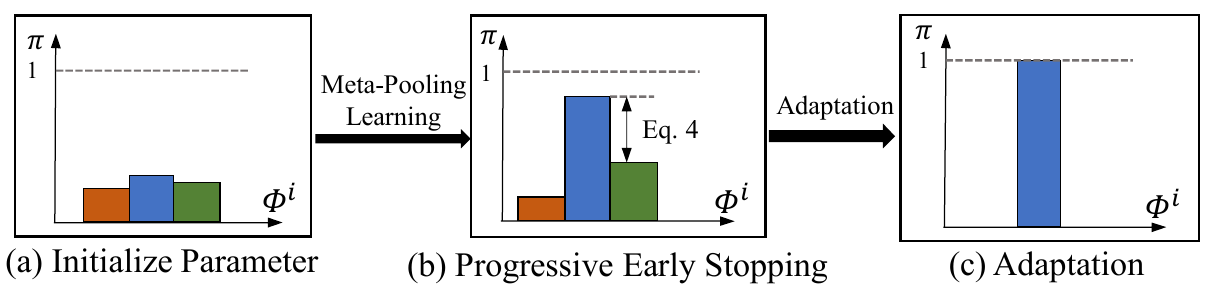}
  \vspace{-3mm}
  \caption{Overview of the Progressive Meta-Pooling Learning. (a) Initialize the meta-parameter $\alpha^i$. (b) Progressive early stopping when Eq.\ref{dominating_condition} is satisfied. (c) Adapt the learned spatial enhancers.}
  \label{fig:progressive}
  \vspace{-3mm}
\end{figure}

\begin{figure*}
  \centering
  \includegraphics[width=0.9\textwidth]{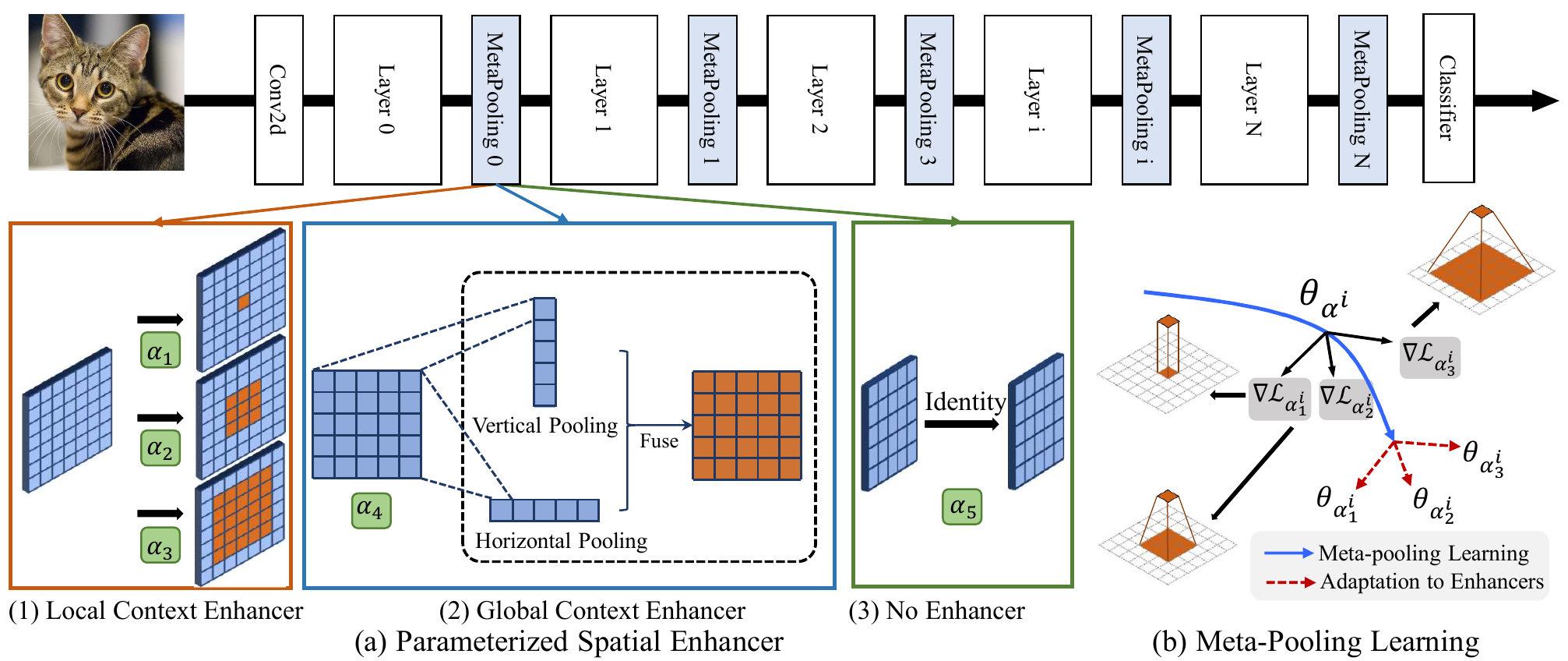}
  \vspace{-3mm}
  \caption{Overview of the proposed Meta-Pooling. We insert the independent Meta-Pooling after each layer of a lightweight network. (a) Parameterized spatial enhancer encompasses three receptive field enhancers, a.k.a. local,  global, and no enhancer. The meta-parameter $\alpha$ is adopted to measure the importance of different enhancers. (b) Diagram of our meta-pooling learning algorithm, which optimizes for a representation $\theta_{\alpha^i}$ that can adapt to the enhancers of the $i$-th layer.}
  \label{fig:main}
  \vspace{-4mm}
\end{figure*}

This study is to determine a suitable receptive field for each layer of a lightweight network in a data-driven way. To make the receptive field learnable, we resort to meta-learning approaches. Meta-learning aims to learn knowledge shared across multiple tasks, and quickly adapt the model to new unseen tasks. Model-Agnostic Meta-Learning\cite{finn2017model} (MAML) has been spotlighted due to its simplicity and generality. To date, otsuzuki \cite{Otsuzuki2021MetalearningOP} proposes to use meta-learning to learn $L_p$ pooing as well as a binary mask to determine the kernel shape of pooling layers for character recognition, but neglects to investigate receptive fields. We apply the concept of MAML to receptive field search by assuming the multiple tasks in MAML as learning for multiple receptive fields. MAML minimizes the expectation of loss across tasks while we minimize the expectation of loss across various receptive fields.

In this paper, we present a meta-learning-based approach called Meta-Pooling to make the receptive field of lightweight networks learnable and enhance the feature discriminability and robustness. As shown in Fig.\ref{fig:main}(a), our proposed Meta-Pooling consists of a parameterized spatial enhancer, including a local context enhancer, a global context enhancer, and no enhancer, which provides the lightweight model with diverse sizes of receptive fields. We employ a Progressive Meta-Pooling Learning (PMPL) strategy with early stopping to learn the distribution of parameterized spatial enhancers, as shown in Fig.\ref{fig:main}(b) and Fig.\ref{fig:progressive}.

Our contributions are summarized as follows: 

(1) We present a Meta-Pooling framework to learn the versatile receptive fields for the lightweight network, which is composed of parameterized spatial enhancers. 

(2) We propose Progressive Meta-Pooling Learning with an early stopping strategy that allows the model to gradually determine the receptive fields. 

(3) We demonstrate that Meta-Pooling outperforms the baseline by 2.3\% and exhibits a remarkable performance compared with pooling and attention-based methods.

\vspace{-0.5cm}

\section{Methodology}

We describe the parameterized spatial enhancers in Sec.\ref{spatial_enhancer}, which provides the lightweight model variety of receptive field patterns. We then introduce a meta-learning-based approach to make the  receptive field learnable with progressive early stopping in Sec.\ref{meta-pooling-learning}.

\subsection{Parameterized Spatial Enhancer}\label{spatial_enhancer}

Parameterized spatial enhancer can provide a variety of receptive fields for the lightweight models. As shown in Fig.\ref{fig:main}, it includes a global context enhancer, a local context enhancer, and no enhancer. The global and local context enhancer can enlarge the receptive field while no enhancer represents the skip connection and can maintain the receptive field.

\noindent\textbf{Global Context Enhancer.} Global Context Enhancer enables the lightweight model to capture an informative long-range context from horizontal and vertical directions. To capture long-range interactions spatially with less computational burden, we factorize the global pooling from the vertical and horizontal directions, inspired by  CoordAttention\cite{Hou2021CoordinateAF} and Strip Pooling\cite{Hou2020StripPR}. We present a spatial separable pooling operation, which is capable to extract global context economically. Specifically, given the input data $X \in \mathbb{R}^{C \times H \times W}$, two spatial extents of pooling kernels $(H,1)$ or $(1,W)$ are adopted to encode horizontal and vertical direction, respectively. The information encoded from the horizontal directions can be formulated as: $z_c^h(h)=\frac{1}{W} \sum_{0 \leq i<W} x_c(h, i)$. Similarly, the information aggregated from the vertical direction can be written as: $z_c^w(w)=\frac{1}{H} \sum_{0 \leq j<H} x_c(j, w)$. Then, we use point-wise convolution to refine the knowledge from the direction-aware feature map and apply Sigmoid to get the spatial activation. The final output of spatial separable pooling is obtained by refining the original feature map with the spatial activation.

\SetInd{0.5em}{0.5em}
\begin{algorithm}[t]
	\caption{Progressive Meta-Pooling Learning.}
	\label{alg:algorithm1}
	\KwIn{layer depth $L$, model weight $\theta$, meta-parameter $\{\alpha^i\}^L$, learning rate $\zeta$ for the inner loop and $\eta$ for the outer loop, early stopping threshold $\rho$, loss on meta dataset $\mathcal{L}_{mt}$, loss on training dataset $\mathcal{L}_{tr}$}
	\While{\textnormal{not all fixed}}{

        Update meta-parmaeters $\alpha$ by $\theta^{'} \leftarrow \theta-\zeta \nabla_\theta \mathcal{L}_{tr}\left(\theta, \alpha\right)$; $\alpha \leftarrow \alpha - \eta \nabla_\alpha \mathcal{L}_{mt}( \theta^{'}, \alpha)$;
        
        Update model weight $\theta$ by $\theta \leftarrow \theta - \zeta \nabla_\theta\mathcal{L}_{tr}\left(\theta, \alpha\right)$
		
		\ForEach{$l \leftarrow 0$ to $L-1$}
		{ 
		    $\pi^l \leftarrow \textnormal{Softmax}(\alpha^l)$
		    
		    $\pi_k^l \leftarrow \max\{\pi^l\}$
		    
		    diff $\leftarrow$ 0
		    
		    \ForEach{$m \leftarrow 0$ to $|\pi|$}
		    {
                $\textnormal{diff} \leftarrow \min \left(\textnormal{diff}, \pi^l_k-\pi^l_m \right)$
    		}
    		\If{$\textnormal{diff} \geq \rho$}{
    		     Fix the distribution by $\arg\max\pi^l$;
		    }
		}
	}
\vspace{-0.2cm}
\end{algorithm}

\noindent\textbf{Local Context Enhancer.} Local Context Enhancer compensate for the lack of local context by introducing a slightly larger kernel. It includes a series of pooling operations with a large kernel, including average pooling and max pooling. In order to provide a more hierarchical receptive field expansion option to the lightweight model, the max pooling and average pooling with diverse kernel sizes $k=\{3,5,7\}$ are utilized to gather local contexts. 

\noindent\textbf{Parameterized Spatial Enhancer.} To adaptively determine the receptive field, we parameterize the above spatial enhancers to form the Meta-Pooling and make it learnable. Specifically, the meta-parameters $\alpha^i = \{\alpha_1^i,\alpha_2^i, .., \alpha_n^i \}$ are introduced to measure the importance of different spatial enhancers $\Phi^i=\{\beta^i_1,\beta^i_2,...,\beta^i_n\}$, where $\alpha_n^i$ means the $n$-th parameter of the $i$-th layer and $\beta_j^i$ means the $j$-th receptive field expansion operations of the $i$-th layer. By meta-parameters, we relax the categorical choice of a particular receptive field and make the learning process end-to-end.

\subsection{Progressive Meta-Pooling Learning}\label{meta-pooling-learning}

We introduce a Progressive Meta-Pooling Learning (PMPL) algorithm to optimize the parameterized spatial enhancers. Moreover, a progressive early stopping mechanism is adopted to accelerate the training process. The PMPL allows the layers of the model at various depths $d=\{d^0, d^1, .., d^n\}$ to automatically measure the importance of different spatial enhancers. 

Given the parameterized spatial enhancers $\Phi^i$, we employ categorical distribution $\beta \sim p(\beta |\pi)$ with probabilities $\pi^i$ for each depth $d^i$. We apply continuous relaxation to meta-parameter $\alpha^i$, and the probabilities are computed as a Softmax over $\alpha^i$ defining the importance for the corresponding spatial enhancer $\Phi^i$ of the $i$-th layer by:
{\setlength\abovedisplayskip{0.2cm}
\setlength\belowdisplayskip{0.2cm}
\begin{equation}
\begin{split}
    \bar{\beta^i}(x)=\sum_{k=0}^{|\Phi^i|} \pi^i_k \beta_k(x),
    \pi_{j}^i=\frac{\exp \left(\alpha^i_{j}\right)}{\sum_{k=0}^{|\Phi|} \exp \left(\alpha^i_{k}\right)}
\end{split}
\end{equation}}
\noindent where $x$ and $\bar{\beta^i}$ are the input and mixed output of the $i$-th layer. In this way,  learning the receptive field for a lightweight model can be viewed as MAML of each spatial enhancer. Given $\mathcal{T}$ as the given dataset, our meta-pooling learning approach is formulated as follows:
\begin{gather}
    \alpha^i = \arg\min_{\alpha^i} \mathbb{E}_{\beta \sim p(\beta|\pi)}\left[\mathcal{L}_{\tau}\left(f_{\theta^*(\alpha^i)}^{i+1:}, \alpha^i \right)\right] \notag \\
    s.t. \quad \tau = p\left(f_{\theta^*}^{0:i-1}\left(x\right), y | \{x, y\} \sim \mathcal{T}\right) \\
    \theta^*(\alpha^i) = \arg\min_{\theta^*} \mathbb{E}[\mathcal{L}_{\mathcal{T}}(\theta, \alpha^i)] \notag
\end{gather}
\noindent where $\theta$ denotes the weights of the lightweight model and $f^{i:j}$ is a composition from $i$-th layer to $j$-th layer. $f^{0:i-1}(x)$ means the intermediate distribution from the last layer, and the distribution is determined by all the previous layers from $p(f)$. We approximate the bi-level objective through first-order model-agnostic meta-learning as follows:
{\setlength\abovedisplayskip{0.2cm}
\setlength\belowdisplayskip{0.2cm}
\begin{equation}
    \begin{split}
    \theta^{'} \leftarrow \theta-\zeta \nabla_\theta \mathcal{L}_{tr}\left(\theta, \alpha^i\right) \\
    \alpha^i \leftarrow \alpha^i - \eta \sum_i \nabla_{\alpha^i} \mathcal{L}_{mt}\left(\theta^{'}, \alpha^i\right)
\end{split}
\end{equation}}
\noindent where $\zeta$ and $\eta$ are the learning rate of $\theta$ and $\alpha^i$. Denoted by $\mathcal{L}_{mt}$ and $\mathcal{L}_{tr}$ is the loss on the meta dataset and training dataset, respectively. The meta dataset is created by separating the original training dataset.

To reduce the computational cost, we propose the progressive early stopping mechanism to progressively discrete the lightweight network by layer through the distribution of parameter $\pi^i$, as shown in Fig.\ref{fig:progressive}. Specifically, a threshold $\rho$ is set to determine whether to stop searching for the $i$-th layer and the condition is as follows:
\begin{gather}
    \min \left\{\pi_{k}^i-\pi_{m}^i, \forall m \neq k | \pi_{k}^i=\max \left\{\boldsymbol{\pi^{i}}\right\}\right\} \geq \rho \label{dominating_condition}
\end{gather}

Once the above condition is met on any layer, the receptive field operation with the highest probability would be fixed, whose parameters would stop updating in the subsequent training. After all of the meta-parameters are fixed, we adapt the learned spatial enhancer to the original task and train the lightweight network from scratch. The iterative procedure is outlined in Alg.\ref{alg:algorithm1}.

\begin{table}[]
\caption{Comparision with pooling-based methods on ImageNet dataset. All methods are based on MobileNetV2~\cite{Sandler2018MobileNetV2IR}}. 
\centering 
\begin{tabular}{lcc}
\hline
             & Top1 acc(\%) & Top5 acc(\%) \\ \hline
Skip         & 71.34                        & 90.30    \\
MaxPool      & 71.35                        & 90.18    \\
AveragePool  & 71.68                        & 90.28    \\ \hline
BlurPool\cite{Zhang2019MakingCN} & 69.42                        & 88.74    \\
DPP\cite{Saeedan2018DetailPreservingPI}  & 70.15                        & 89.47    \\
GatedPool\cite{Lee2016GeneralizingPF}  & 71.06                        & 90.10     \\
GaussPool\cite{Kobayashi2019GaussianBasedPF}    & 72.87                        & 91.08    \\
LiftDownPool \cite{Zhao2021LiftPoolBC} & 73.91                        & 91.78    \\
Meta-Pooling(Ours) & \textbf{74.60}                  & \textbf{91.81}    \\ \hline
\end{tabular}\label{tab:pooling-based}
\end{table}

\begin{table}[]
\caption{Comparison of attention-based methods under width multipliers of $\{1.0,0.75,0.5\}$.}
    \resizebox{\linewidth}{!}{
    \begin{tabular}{lccl}
    \hline
                    & \multicolumn{1}{l}{Param.(M)} & \multicolumn{1}{l}{M-Adds(M)} & \multicolumn{1}{l}{Top-1 Acc(\%)} \\ \hline
    MobileNetV2     & 3.50                          & 300                           & $72.3$                       \\
    +SE\cite{Hu2020SqueezeandExcitationN} & 3.89                          & 300                           & $73.5_{+1.2}$                \\
    +CBAM\cite{Woo2018CBAMCB}            & 3.89                          & 300                           & $73.6_{+1.3}$                 \\
    +Meta-Pooling(Ours)            & 3.91                          & 320                           & $74.6_{+2.3}$                \\ 
    \hline
    MobileNetV2-0.75 & 2.5                          & 200                           & $69.9$                            \\
    +SE\cite{Hu2020SqueezeandExcitationN} & 2.86                          & 210                           & $71.5_{+1.6}$                      \\
    +CBAM\cite{Woo2018CBAMCB}            & 2.86                          & 210                           & $71.5_{+1.6}$                      \\
    +Meta-Pooling(Ours)            &    2.86                       &    210                      &    $71.8_{+1.8}$                        \\ 
    \hline
    MobileNetV2-0.5  & 2.0                          & 100                           & $65.4$                          \\
    +SE\cite{Hu2020SqueezeandExcitationN} & 2.1                           & 100                           & $66.4_{+1.0}$                      \\
    +CBAM\cite{Woo2018CBAMCB}            & 2.1                           & 100                           & $66.4_{+1.0}$                      \\
    +Meta-Pooling(Ours)            & 2.1                           & 100                           & $66.9_{+1.5}$                              \\ \hline
    \end{tabular}\label{weight_multiplier}
    }
    \vspace{-4mm}
\end{table}

\section{Experiments}

We first introduce the experiment setup in Sec.\ref{setup} and then carry out various experiments based on MobileNetV2 in Sec.\ref{main_results}. Finally, we perform an ablation study to verify the contribution of our proposed Meta-Pooling in Sec.\ref{abla}.

\subsection{Experiment Setup}\label{setup}

Our experiments are implemented with PyTorch. The SGD optimizer with a momentum of 0.9 and weight decay of 4e-5 is adopted to train the lightweight models, and the Adam optimizer is utilized to train the meta-parameters with a weight decay of 1e-3. The cosine learning schedule is adopted for SGD optimizer. We use 4 NVIDIA GPUs for training and the batch size is set to 128. The threshold $\rho$ of progressive early stopping is 0.2. We first train for 30 epochs by PMPL and retrain for 200 epochs when the receptive fields of all layers are fixed. The learning rate of $\zeta$ is 0.05 and $\eta$ is 3e-4. We use the same data augmentation as in MobileNetV2\cite{Sandler2018MobileNetV2IR}. All of the experiments are conducted on ImageNet dataset\cite{krizhevsky2012imagenet}.

\subsection{Results}\label{main_results}

Using the progressive early stopping strategy enables the model to converge quickly within 30 epochs. We notice that deeper layers tend to employ Global Context Enhancer, which reveals the lack of receptive field in lightweight networks. We further compared Meta-Pooling with other methods.

\noindent \textbf{Pooling-based Methods.} We compared Meta-Pooling with a range of pooling-based approaches\cite{Zhang2019MakingCN,Saeedan2018DetailPreservingPI,Lee2016GeneralizingPF,Kobayashi2019GaussianBasedPF,Zhao2021LiftPoolBC}, as shown in the Tab.\ref{tab:pooling-based}. Meta-Pooling achieves 3\% higher accuracy than max pooling and average pooling. The pooling-based operations can well preserve detailed information and position shift-invariant but fail to explore larger receptive fields.

\noindent \textbf{Attention-based Methods.} We compare our Meta-Pooling with other lightweight attention methods, such as SE\cite{Hu2020SqueezeandExcitationN}, CBAM\cite{Woo2018CBAMCB}, etc. As shown in Tab.\ref{pooling_results}, the SE and CBAM improve the performance by 1\%, while our Meta-Pooling shows acceptable improvements by 2.3\%, which is comparable to CoordAttention~\cite{Hou2021CoordinateAF}.

\noindent \textbf{Different Width Multiplier.} We adopt three weight multipliers, which are $\{1.0, 0.75, 0.5\}$. As shown in Tab.\ref{weight_multiplier}, Meta-Pooling achieves better trade-off compared with SE and CBAM under different width multipliers. However, we found that the diminishing improvement brought by Meta-Pooling as the model becomes smaller, which may be due to the difficulty of optimizing the large receptive field for smaller models. 

\subsection{Ablation Study}\label{abla}

The ablation study of the spatial enhancers and learning strategy is shown in Tab.\ref{tab:ablation}, which exhibits the effectiveness of our approach. We observe that a larger receptive field can indeed improve the performance of lightweight networks, which . For example, simply using global can achieve a 2\% improvement over the baseline. Besides, our proposed PMPL strategy has an improvement of 1.2\% compared with the random selection of spatial enhancers.

\begin{table}[]
\small
\centering
\caption{Ablation study of enhancers and training strategy. "No" refers to skip connection. "Local" refers to an average pooling layer with a kernel size of 3. "Global" refers to spatial separable pooling. "Random" refers to a random selection from three enhancers. "PMPL" refers to the progressive meta-pooling learning strategy with early stopping.}
\begin{tabular}{ccccc|l}
    		\hline
    		No   &Local        & Global       & Random       & PMPL  &      Acc(\%) \\ \hline
    		$\checkmark$ & -            & -            & -            & -            & 72.3          \\
    		-            &$\checkmark$ & -            & -            & -            & 71.6$_{-0.7}$  \\
    		-            &-            & $\checkmark$ & -            & -            & 74.3$_{+2.0}$ \\
            $\checkmark$ &$\checkmark$ & $\checkmark$ & $\checkmark$ & -            & 73.4$_{+1.1}$ \\
    		$\checkmark$ &$\checkmark$ & $\checkmark$ & -            & $\checkmark$ & 74.6$_{+2.3}$ \\ \hline
\end{tabular}\label{tab:ablation}
\end{table}
\vspace{-4mm}

\begin{table}[]
\centering
\caption{Comparison with attention-based methods on ImageNet dataset.}
\resizebox{\linewidth}{!}{
    \begin{tabular}{lcccc}
    \hline
            & Params & GFLOPs & Top-1(\%) & Top-5 (\%) \\ \hline
    MobileNetV2\cite{Sandler2018MobileNetV2IR}  & 3.50M      & 0.32   & 72.3     & 90.0      \\
    +SE\cite{Hu2020SqueezeandExcitationN} & 3.56M      & 0.32   & 73.5    & 90.5      \\
    +CBAM\cite{Woo2018CBAMCB}         & 3.57M      & 0.32   & 73.6    & 90.5      \\
    +SK\cite{Li2019SelectiveKN}           & 5.28M      & 0.35   & 74.1    & 91.8      \\
    +CoordAttention\cite{Hou2021CoordinateAF} & 3.95M   & 0.33    & 74.3    & \textbf{91.9}      \\
    +Meta-Pooling(Ours)         & 3.91M      & 0.32    & \textbf{74.6}    & 91.8 \\ \hline
    \end{tabular}\label{pooling_results}
}
\vspace{-3mm}
\end{table}

\section{Conclusion}

We present Meta-Pooling to make the receptive field learnable for lightweight networks. By introducing the parameterized spatial enhancer, the Meta-Pooling can model receptive fields of diverse sizes. The Progressive Meta-Pooling Learning strategy with early stopping is employed to optimize the parameterized spatial enhancer. The results demonstrate that our Meta-Pooling makes up for the lack of receptive field and achieve remarkable performance. 

\vfill\pagebreak

\bibliographystyle{IEEEbib}
\bibliography{strings,refs}

\end{document}